\documentclass[sigconf]{acmart}

\usepackage{wrapfig}
\usepackage{booktabs} 
\usepackage{soul}
\usepackage{paralist}
\usepackage{amsmath}
\usepackage{multirow}
\usepackage{tabularx}
\usepackage{balance}
\usepackage{booktabs}
\usepackage{subcaption}
\usepackage{enumitem}
\usepackage{hyperref}
\usepackage[skip=7pt]{caption}
\usepackage[ruled,vlined,linesnumbered]{algorithm2e}
\SetArgSty{textnormal}
\setcopyright{rightsretained}
\usepackage{etoolbox}

\apptocmd{\thebibliography}{\small}{}{}
\def\sharedaffiliation{
	\begin{tabular}{c}}
	
\begin{document}

\title{Linked Causal Variational Autoencoder for \\ Inferring Paired Spillover Effects}

\author{Vineeth Rakesh}

\authornote{Both authors contributed equally to the paper}
\affiliation{%
	\institution{Arizona State University}
}
\email{vrakesh@asu.edu}

\author{Ruocheng Guo}
\authornotemark[1]
\affiliation{%
	\institution{Arizona State University}
}
\email{rguo12@asu.edu}

\author{Raha Moraffah}
\affiliation{%
	\institution{Arizona State University}
}
\email{rmoraffa@asu.edu}

\author{Nitin Agarwal}
\affiliation{%
	\institution{University of Arkansas at Little Rock}
}
\email{nxagarwal@ualr.edu}

\author{Huan Liu}
\affiliation{%
	\institution{Arizona State University}
}
\email{huan.liu@asu.edu}

\begin{abstract}

Modeling spillover effects from observational data is an important problem in economics, business, and other fields of research.
It helps us infer the causality between two seemingly unrelated set of events. For example, if consumer spending in the United States declines, it has spillover effects on economies that depend on the U.S. as their largest export market. In this paper, we aim to infer the causation that results in spillover effects between pairs of entities (or units); we call this effect as \textit{paired spillover}. To achieve this, we leverage the recent developments in variational inference and deep learning techniques to propose a generative model called Linked Causal Variational Autoencoder (LCVA). Similar to variational autoencoders (VAE), LCVA incorporates an encoder neural network to learn the latent attributes and a decoder network to reconstruct the inputs. However, unlike VAE, LCVA treats the \textit{latent attributes as confounders that are assumed to affect both the treatment and the outcome of units}. Specifically, given a pair of units $u$ and $\bar{u}$, their individual treatment and outcomes, the encoder network of LCVA samples the confounders by conditioning on the observed covariates of $u$, the treatments of both $u$ and $\bar{u}$ and the outcome of $u$. Once inferred, the latent attributes (or confounders) of $u$ captures the spillover effect of $\bar{u}$ on $u$. Using a network of users from job training dataset (LaLonde (1986)) and co-purchase dataset from Amazon e-commerce domain, we show that LCVA is significantly more robust than existing methods in capturing spillover effects.
\vspace{-2mm}
\end{abstract}
\vspace{-2mm}
\vspace{-2mm}
\vspace{-2mm}
\vspace{-2mm}
\vspace{-2mm}	

\copyrightyear{2018}
\acmYear{2018}
\setcopyright{acmcopyright}
\acmConference[CIKM '18]{The 27th ACM International Conference on
	Information and Knowledge Management}{October 22--26, 2018}{Torino,
	Italy}
\acmBooktitle{The 27th ACM International Conference on Information and
	Knowledge Management (CIKM '18), October 22--26, 2018, Torino, Italy}
\acmPrice{15.00}
\acmDOI{10.1145/3269206.3269267}
\acmISBN{978-1-4503-6014-2/18/10}

\maketitle
\vspace{-2mm}
\section{Introduction}

\begin{figure}[!tb]
	\includegraphics[scale=0.31]{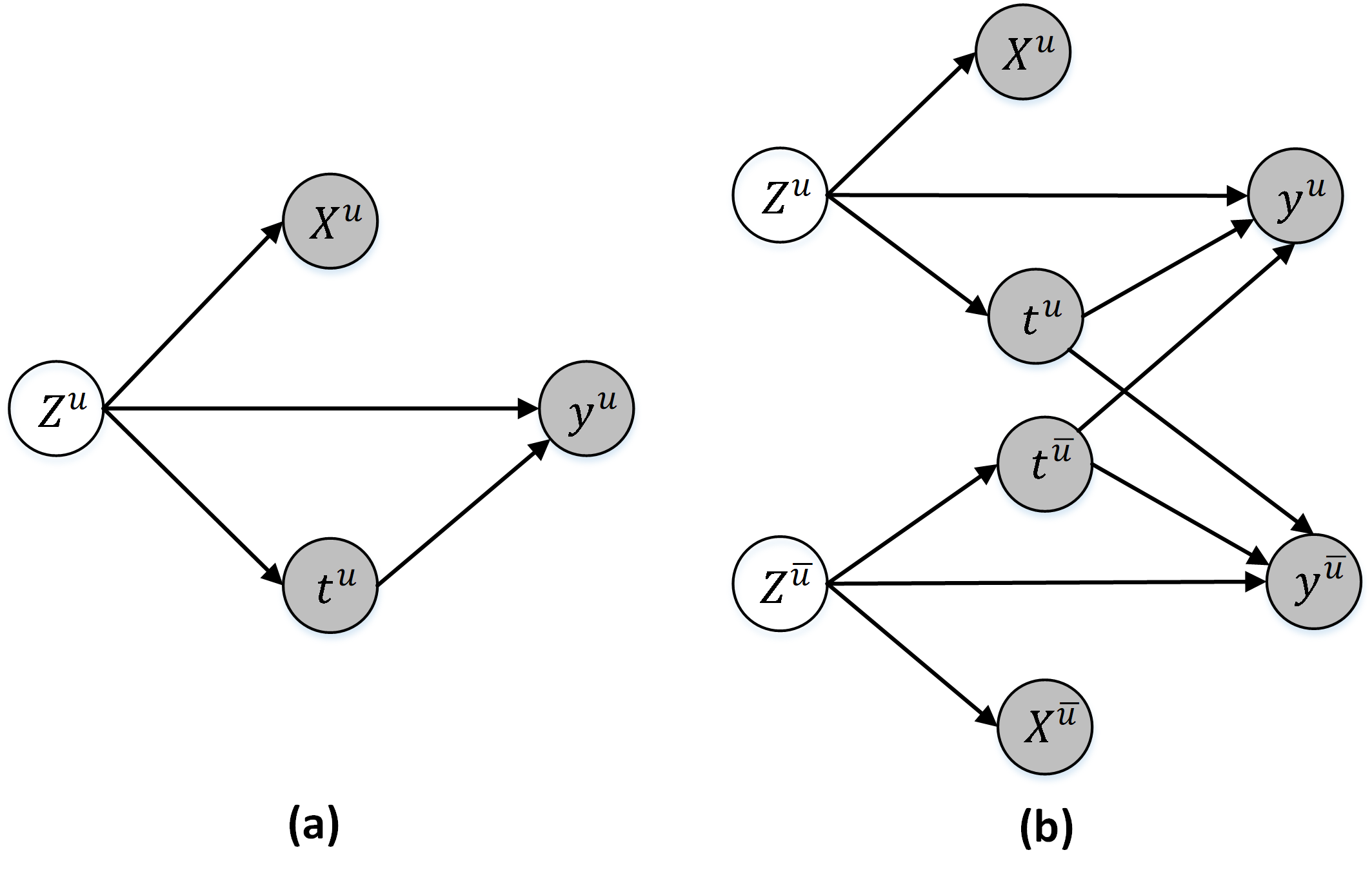}
	\caption{Graphical structure of (a) conventional and (b) linked causal effect model. Here, $(u,\bar{u})$ are a pair of units, $X^{*}$, $t^{*}$ and $y^{*}$ are the observed covariates (or proxies) of units, the treatment and the outcome where $*$ indicates either $u$ or $\bar{u}$.}
	\label{fig:spillover_example}
\end{figure}
\vspace{-1mm}
Interference in causal inference is when the outcome of a unit is not only influenced by its own treatment and covariates but also by those of other units. In economics, this phenomenon is known as the \textit{spillover effect}~\cite{sobel2006randomized}. Understanding spillover effects is extremely important to answer questions such as: Will eradicating pests from one farm cause them to move to nearby farms? Will the popularity of one product drive the sales of another product? Will the introduction of genetically modified crops result in the contamination of neighboring organic crops?
%
In this paper, our goal is to estimate the causal effect of a certain treatment (e.g. positive review) on a specific outcome (e.g. product sales) when there exist spillover effects between pairs of entities. For instance, positive reviews for XBOX consoles could result in increased sales of other XBOX accessories such as controllers and games. We term this effect as \textit{paired spillover}.

Research on causal Machine Learning has gained significant attention in recent years. This can be attributed to their empirical success in estimation of unknown functions without strong model specification. 
For instance, in~\cite{hill2011bayesian}, the authors show that direct regression adjustment with Bayesian Additive Regression Trees (BART) leads to promising estimation of individual treatment effects. \cite{johansson2016learning} propose a model for counterfactual inference by bringing together ideas from representation learning and domain adaptation. \cite{shalit2016estimating} and \cite{louizos2017causal} utilize deep learning techniques to infer individual treatment effects and counterfactual outcome. Despite their novel approach, none of these studies account for spillover effects. 

To overcome the aforementioned problem, we leverage the latest developments in variational deep learning techniques \cite{kingma2013auto,louizos2017causal} to propose a model called Linked Causal Variational Autoencoder (LCVA). The framework of LCVA is based on variational autoencoder (VAE) \cite{kingma2013auto}, which incorporates an encoder neural network to learn the latent attributes and a decoder network to reconstruct the inputs. Nonetheless, unlike VAE, LCVA adopts the inferential process of a causal variant of VAE called \textit{causal effect variational autoencoder} (CEVAE) \cite{louizos2017causal}. The graphical structure shown in Figure \ref{fig:spillover_example} (a) explains the generative principle of CEVAE. Here, $X$ is a set of covariates of a unit $u$, $t$ is the treatment, $y$ is the outcome and $Z$ is the confounder. CEVAE treats the confounder as a latent variable and conditions the outcome and treatment on the hidden confounder. We propose an extended framework in Figure \ref{fig:spillover_example} (b) where $u$ and $\bar{u}$ are two units and the spillover effect is modeled by allowing the treatment of unit $u$ (i.e., $t^{u}$) to influence the outcome of unit $\bar{u}$ (i.e., $y^{\bar{u}}$) and vice versa. Our objective is to learn the confounders $Z^{u}$ and $Z^{\bar{u}}$.  A common practice of inferring such counfounders is to use \textit{proxy variables} \cite{maddala1992introduction,montgomery2000measuring}. For example, in the study of causal effect of job training on annual income, we cannot measure every attribute that influences the earnings of an individual, but we might be able to get a proxy for it through a set of accessible variables such as zip code and job type. There are several ways to use these proxies to estimate $Z$. Louizos et al.~\cite{louizos2017causal} showed that one of the effective ways is to directly condition the proxy on $Z$ and infer $Z$ using approximate maximum-likelihood based methods. Therefore, we adopt the same technique by allowing the covariates $X^{u}$ and $X^{\bar{u}}$ to act as proxies of confounders $Z^{u}$ and $Z^{\bar{u}}$, respectively.

%

To evaluate the proposed model, it is important to consider datasets where the outcome of units is influenced by some form of spillover effect. Ideally, the dataset needs to have the following properties:
\begin{inparaenum}[(1)]
	\item network information in the form of links between units and 
	\item the counterfactual outcome of individual units. 
\end{inparaenum}
Unfortunately, most existing datasets do not contain both these properties together. Therefore, we modify the following real word datasets by filling-in the missing information: 
\begin{inparaenum}[(a)]
	\item the job training dataset~\cite{lalonde1986evaluating,smith2005does} and
	\item the co-purchase dataset from Amazon e-commerce domain \cite{mcauley2015inferring}.
\end{inparaenum}
In particular, the job training dataset does not include any network information; consequently, similar to~\cite{li2018unsupervised}, we create a K-NN graph based on the covariates of units to connect similar individuals.
This can be justified by the theory of Homophily \cite{mcpherson2001birds,zafarani2014social,shakarian2015diffusion} which states that \textit{birds of a feather flock together}. Contrary to the job training dataset, the co-purchase dataset from Amazon does contain the network information, which specifically states whether an item is a substitute or a complement of another item \cite{mcauley2015inferring}.
Nonetheless, it does not have the counterfactual outcome.  In our case, the counterfactual is the sales of a product if it had no reviews. We synthesize this using a matching technique that was introduced by Kuang et al. \cite{kuang2017estimating}. To the best of our knowledge, this is the very first deep variational inference framework that is specifically designed to infer the causality of spillover effects between pairs of units. The major contributions of this paper are detailed as follows:
\vspace{-1mm}
\begin{itemize}[leftmargin=*]
	\itemsep0em
	\item We propose a model called linked causal variational autoencoder (LCVA) that captures the spillover effect between pairs of units. Specifically, given a pair of units $u$ and $\bar{u}$, their individual treatment and outcomes, the encoder network of LCVA samples the confounders by conditioning on the observed covariates of $u$, the treatments of both $u$ and $\bar{u}$ and the outcome of $u$. We introduce two new datasets: a job training dataset \cite{lalonde1986evaluating,smith2005does} that is augmented with synthesized network information and an Amazon dataset \cite{mcauley2015inferring} that is augmented with counterfactual outcomes.
\item Using a rigorous series of experiments, we show that LCVA is extremely effective in capturing spillover effects between units. It also beats existing methods on various metrics across all datasets.
\end{itemize}


\begin{figure}[!tb]
	\includegraphics[width=0.9\columnwidth]{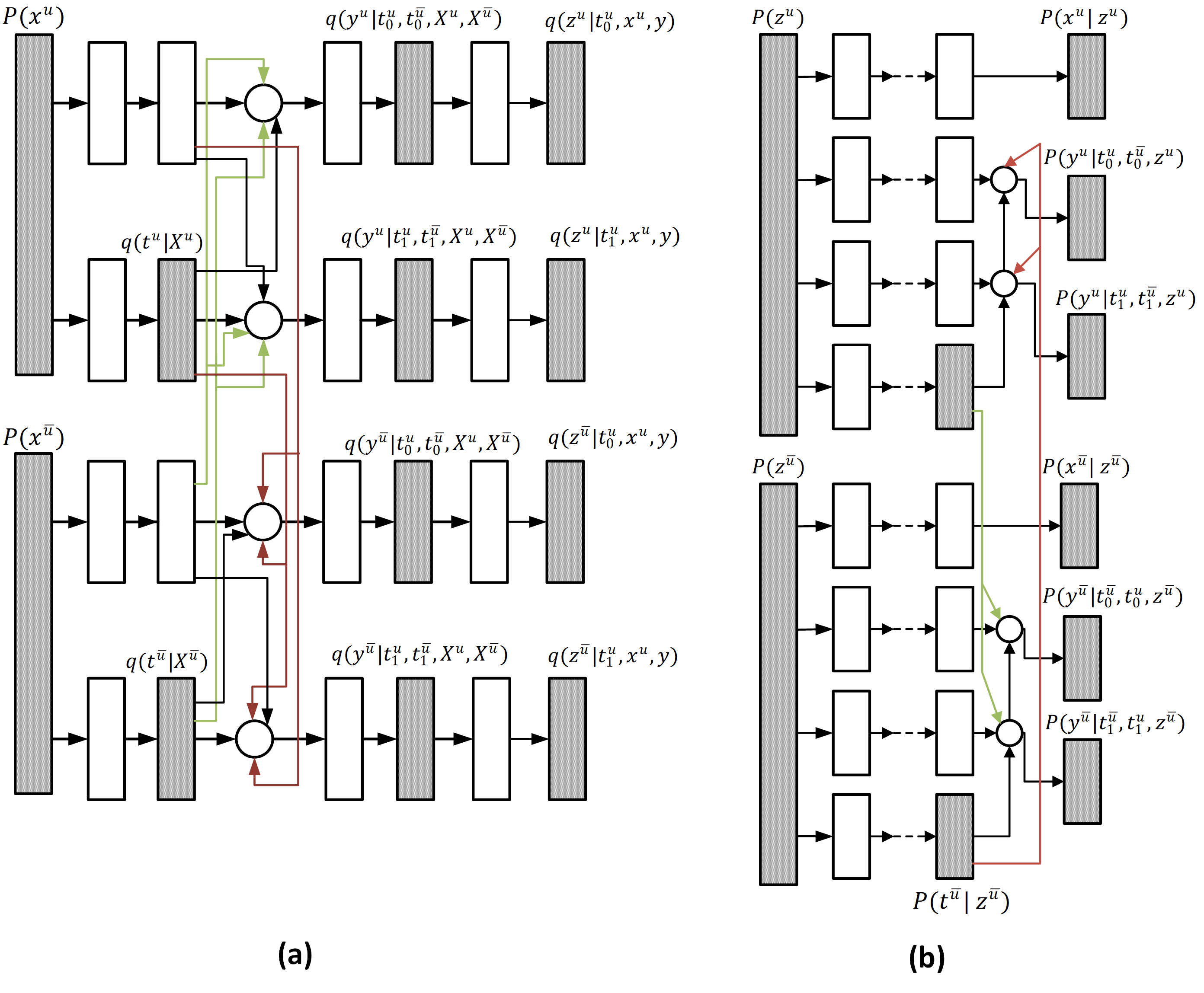}
	\caption{The architecture of LCVA model; (a) is the encoder part and (b) is the decoder. The red lines indicate the effect of unit $u$'s covariates and treatment on the network of $\bar{u}$ and the red lines indicate the same from $\bar{u}$ to $u$.}.
	\label{fig:LCVA}
\end{figure}

\vspace{-4mm}
\section{The Proposed LCVA Model}
\vspace{-1mm}
\setlength{\abovedisplayskip}{0pt}
As explained in the previous section, our objective is to infer the latent confounders $Z^{*}$, where $*$ indicates $u$ or $\bar{u}$. To achieve this, we assume that the latent confounder $z^{*}$ can be sampled from the observed variables $x^{*},t^{*},y^{*}$ and the treatment $t^{\hat{*}}$, where $\hat{*}$ indicates $\bar{u}$ if $* = u$ and $u$ if $* = \bar{u}$, $x$ indicates the covariates of a single unit and $z$ indicates the confounder of a single unit. $z^{*}$ is inferred using the proposed linked causal variational autoencoder (LCVA) that is depicted in Figure \ref{fig:LCVA} . Here, the white nodes correspond to parametrized deterministic neural network, the grey nodes correspond to drawing samples from the respective distribution and the white circles correspond to switching path according to the treatment $t$. Learning the latent variable $z$ is typical of any variational inference technique,  where the objective is to optimize the KL-divergence between the true posterior $p(z^{*}|\cdot)$ and the variational distribution $q(z^{*}|\cdot)$. The sampling of $q(z^{*}|\cdot)$ is similar to the recently proposed CEVAE \cite{louizos2017causal}; however, unlike CEVAE, our goal is to capture the paired spillover effect. Therefore, in Figure $q(z^{*}|\cdot)$ (where $*$ = $u$ or $\bar{u}$) is sampled not only based on its individual covariates $x^{*}$, treatments $t^{*}$, and factual outcome $y^{*}$, but also on $t^{\bar{u}}$ and $x^{\bar{u}}$. The green and yellow lines in Figure \ref{fig:LCVA} (a) signify this dependency.

\vspace{-2mm}
\subsection{Parameter Inference}
\vspace{-1mm}
We begin by defining the evidence lower bound (ELBO) of LCVA for unit $u$ as follows:
\vspace{-0mm}
\begin{align}
\mathcal{L} &= \sum_{i=1}^{N} \mathbb{E}_{q_{\phi}(z_{i}^{u} | x_{i}^{u},x_{i}^{\bar{u}},t_{i}^{u},t_{i}^{\bar{u}},y_{i}^{u})} \bigg[log P_{\theta}(x_{i}^{u}|z_{i}^{u}) + log \, P_{\theta}(t_{i}^{u} | z_{i}^{u}) +  \nonumber \\
& log \, P_{\theta}(y_{i}^{u}|t_{i}^{u},t_{i}^{\bar{u}}, z_{i}^{u}) - log \, q_{\phi}(z_{i}^{u},t_{i}^{u},t_{i}^{\bar{u}},y^{u}| x_{i}^{u},x_{i}^{\bar{u}})\bigg] 
\end{align}
where $N$ is the number of units, $\mathbb{E}_{q(z_{i}^{u}|\cdot)}$ is the expectation w.r.t distribution $q$ and $\phi, \theta$ are the weights of encoder and decoder network respectively. The ELBO of unit $\bar{u}$ remains similar to the above expression; hence, we do not exclusively derive the lower bound for $\bar{u}$. From the encoder's neural network (Figure \ref{fig:LCVA}(a)), one can observe that the approximate posterior $q(z|\cdot)$ factorizes as follows:
\begin{align}
\sum_{i=1}^{N} & q_{\phi}(z_{i}^{u}|t_{i}^{u},t_{i}^{\bar{u}}, x_{i}^{u},x_{i}^{\bar{u}},y^{u}) + log \, q_{\phi}(t_{i}^{u} |x_{i}^{u}) + log \, q_{\phi}(t_{i}^{\bar{u}} | x_{i}^{\bar{u}}) \nonumber \\
& + log \, q_{\phi}(y_{i}^{u} | x_{i}^{u}, x_{i}^{\bar{u}}, t_{i}^{u}, t_{i}^{\bar{u}}) 
\end{align}
we can obtain an unbiased estimate of the ELBO by sampling $z^{u} \sim q_{\phi}$ and use stochastic gradient descent to optimize it. However, we cannot trivially take gradients w.r.t $\phi$. Therefore,  we incorporate the \textit{reparamaterization trick} \cite{kingma2013auto}, to sidestep this issue. $q(z^{u})$ is then approximated by the following expression:

\begin{align}
q(z_{i}^{u}|t_{i}^{u},t_{i}^{\bar{u}}, x_{i}^{u},x_{i}^{\bar{u}},y^{u}) = \prod_{j=1}^{K} \mathcal{N}(\mu_{ij},\sigma^{2}_{i,j}) \nonumber \\
\boldsymbol{\mu_i} = t_i \, (\mu^{u}_{t=0,i} + \mu^{\bar{u}}_{t=0,i}) +  (1-t_i ) \, (\mu^{u}_{t=1,i} +  \mu^{\bar{u}}_{t=1,i}) \nonumber \\
\boldsymbol{\sigma_i} = t_i \, (\sigma^{u}_{t=0,i} + \sigma^{\bar{u}}_{t=0,i}) +  (1-t_i ) \, (\sigma^{u}_{t=1,i} +  \sigma^{\bar{u}}_{t=1,i}) 
\end{align}
where $K$ is the number of latent features, $\mu_{t}$ is the mean of units that received treatment $t$ and $\sigma_t$ indicates the same for variance. As explained earlier, since $z^{u}$ is unobserved, in the inference network, the outcome $y^{u}$ is influenced by both $t^{\bar{u}}$, $x^{\bar{u}}$ due to property of \textit{common cause}. However, once $z^{u}$ is inferred, the outcome $y^{u}$ simply depends on the sampled treatments $t^{u}$ and $t^{\bar{u}}$. Therefore, in the decoder part (Figure \ref{fig:LCVA}(b)), the variables can be generated as follows:
\begin{align}
p(\boldsymbol{x^{u}_{i}}|\boldsymbol{z}^{u}_{i}) = \prod_{j=1}^{N} p(x^{u}_{i,j}|\boldsymbol{z_i}) \\
p(\boldsymbol{t^{u}_{i}}|\boldsymbol{z}^{u}_{i}) = Bernoulli(\sigma(g_{1}(z_i))) 
\end{align}
\begin{align}
p(y^{u}_{i}|t^{u}_{i},t^{\bar{u}}_{i}\boldsymbol{z^{u}_{i}}) = \mathcal{N}(\hat{\mu}_{i},\hat{\sigma}_{i}) \, \, \,\hat{\mu}_{i} = t^{u}_{i}g_2(\boldsymbol{z}^{u}_{i}) + (1-t^{u}_{i})g_3(\boldsymbol{z}^{u}_{i})
\end{align}
where $p(x^{u}_{i,j}|\boldsymbol{z_i})$ is a gaussian distribution, $\sigma(\cdot)$ is the logistic function, and  $g_{\{\cdot\}}$ is a neural network parameterized by weights $\theta$.
%
\vspace{-2mm}
\section{Experiments}
\label{sec:exp}

In this section, we intend to answer the following question: how accurate can the proposed model be in terms of inferring treatment effects?. To see the effectiveness of the proposed model, we compare it with a state-of-the-art model and four other baselines that are widely regarded as classical methods in causal inference. For each model, we carried out a grid search to decide the hyperparameters. 

\noindent\textbf{OLS-1}. A linear regression model $f:[X^u, t^u, t^v]\rightarrow y^u $ is trained with the treatment of linked units considered. 
We can infer the counterfactual outcomes by applying $f$ on $[X^u,1-t^u,t^v]$ for all pair of units $(u,v)\in \mathcal{E}$.

\noindent\textbf{OLS-2}. Given $(u,v)\in\mathcal{E}$, we train two linear regression models $f_1:[X^u, t^v]\rightarrow y^u,\forall u:t^u=1$ and $f_0:[X^u, t^v]\rightarrow y^u,\forall u:t^u=0$.
Then the counterfactual outcomes can be inferred by $\hat{y}_1^u = f_1(X^u, t^v),\forall u:t^u=0$ and $\hat{y}_0^u = f_0(X^u, t^v),\forall u:t^u=1$.

\noindent\textbf{Random forest}. This model is the same as OLS-1 except the function $f$ is replaced by a random forest.

\noindent\textbf{Causal forest}~\cite{wager2017estimation}. This ensemble model consists of a set of causal trees. 
Each causal tree splits the original feature space into leaves and considers the treatments and outcomes of units in a leaf to come from a randomized set of experiments. 

\noindent\textbf{CEVAE}~\cite{louizos2017causal}. Similar to LCVA, CEVAE learns confounders $Z^u$ for each unit but does not take the spillover effect into consideration.

\begin{table}[]
	\small
	\centering
	\caption{Statistics of Job training dataset and co-purchase dataset of positive and negative reviews from Amazon.}
	\label{tab:dataset}
	\begin{tabular}{|c|c|c|c|c|}
		\hline
		\textbf{Name} & \multicolumn{1}{l|}{\textbf{\#Treated}} & \multicolumn{1}{l|}{\textbf{\#Control}} & \textbf{\#Pairs} & \textbf{\#Feature} \\ \hline
		Job & 297 & 2915 & 3512 & 10 \\ \hline
		+ve Amazon & 50K & 10K & 96132 & 300 \\ \hline
		-ve Amazon & 20K & 5K & 28136 & 300 \\ \hline
	\end{tabular}
\end{table}
\vspace{-3mm}
\subsection{Dataset and Evaluation Metrics}
\vspace{-1mm}
\label{subsec:dataset}
%

\noindent\textbf{Job training dataset}: This dataset is used to study the treatment effect of job training on the earning in the year of 1978. In order to de-randomized the data, following~\cite{shalit2016estimating}, we keep the treatment group of LaLonde's study~\cite{lalonde1986evaluating} ($\mathcal{U}_T$) and combine the control group of this study ($\mathcal{U}_C$) and that from the PSID~\cite{smith2005does} ($\mathcal{U}_{C1}$). Additionally, since this dataset does not include any network information, we create a K-NN graph based on the covariates of units to connect similar individuals. Table \ref{tab:dataset} shows the statistics of our dataset.

\noindent {\bf Evaluation metrics}: Due to lack of ground truth for the counterfactual outcomes, we use average treatment effect on the randomized trial subset and policy risk as evaluation metrics \cite{shalit2016estimating}.
The fact that the LaLonde's study is a randomized trial provides ground truth of the ATE for the subset $\mathcal{U}_T\cup\mathcal{U}_C$ by $ATE=\frac{1}{|\mathcal{U}_T|}\sum_{u\in\mathcal{U}_T}y^u_1 - \frac{1}{|\mathcal{U}_C|}\sum_{u\in\mathcal{U}_C}y^u_0$.
Then, the estimated ATE is $ \frac{\sum_{u\in\mathcal{U}_T\cup\mathcal{U}_C}(\hat{y}^u_1-\hat{y}^u_0)}{|\mathcal{U}_T\cup\mathcal{U}_C|}$, where $y^u$ and $\hat{y}^u$ refer to the factual and inferred counterfactual outcome, respectively.
For a treated (or controlled) unit, $\hat{y}_1^u = y_1^u$ ($\hat{y}_0^u = y_0^u$) and $\hat{y}_0^u$ ($\hat{y}_1^u$) is inferred by the models.
%
We report the absolute difference between the estimated ATE and the ground truth: $\epsilon_{ATE} = |ATE-\hat{ATE}|$.
Moreover, following \cite{shalit2016estimating}, we also report the estimated policy risk ($\hat{pr}$) for the randomized trial subset as $\hat{pr} = 1 - (\mathbb{E}[\tilde{y}^u_1|t^u=1,\pi^u=1]p(\pi^u=1)+\mathbb{E}[\tilde{y}^u_0|t^u=0,\pi^u=0]p(\pi=0))$, where $\pi^u = \mathbf{1}(\hat{y}_1^u-\hat{y}_0^u)$ 
and $\tilde{y}^u $ denotes the factual outcome scaled between $[0,1]$.
Intuitively, the weighted sum of the two expectations denotes the expected potential outcome.

\noindent\textbf{Amazon dataset}: For the Amazon dataset \cite{mcauley2015inferring}, we study the causal effect of positive (or negative) reviews on the sales of products.  For our experiments, we choose the co-purchase data from the electronics category and divide the products (or units) into two groups 
\begin{inparaenum}[(1)]
	\item units that have more than three reviews and
	\item units that have less than three reviews.
\end{inparaenum}
The first group is considered as treated (i.e., t=1), while the second is the control group (i.e., t=0). Considering the fact that positive and negative reviews can affect the sales in different ways, we separate the units in treated group into two different datasets:
\begin{inparaenum}[(a)]
	\item units with positive reviews (when average rating > 3) and
	\item units with negative reviews (when average rating < 3).
\end{inparaenum}
To each of these dataset, we add the units from control group to create the final dataset (Table \ref{tab:dataset}). The features of each unit is created by feeding the review text to a doc2vec model to create a vector of 300 latent features per unit. Lastly, we overcome the lack of counterfactual outcomes using a matching technique \cite{kuang2017estimating} to synthesize counterfactuals. To be specific, for a product $u$, the counterfactual outcome is set as the observed sales of the most similar product with an opposite treatment status i.e., $y^u_{1-t^u} = y^v$, where $v = argmin_{v:t^v=1-t^u} ||X^v - X^u||_2^2$.

\noindent{\bf Evaluation metrics}: We use the following metrics for evaluation: Precision in Estimation of Heterogeneous Effect (PEHE) and absolute error on Average Treatment Effect (ATE) as $PEHE = \frac{1}{N}\sum_{u}((y^u_1-y_0^u)-(\hat{y}^u_1-\hat{y}_0^u))^2$, $\epsilon_{ATE} = |ATE-\hat{ATE}|$, where $ATE=\frac{1}{N}\sum_{u}(y_1^u-y_0^u)$ and $\hat{ATE} = \frac{1}{N}\sum_{u}(\hat{y}_1^u-\hat{y}_0^u)$.

\begin{table}[]
	\small
	\centering
	\caption{Results for the job training dataset.}
	\vspace{-1mm}
	\label{tab:ECON_result}
	\begin{tabular}{ccc}
		\hline
		\textbf{Models} & $\epsilon_{ATE}$ & \textbf{Policy Risk} \\ \hline
		OLS1 & 492.51 & 0.87 \\
		OLS2 & 498.05 & 0.86 \\
		RF & 4.05 & 0.84 \\
		CF & 511.68 &  0.93 \\
		CEVAE & 112.46 & 0.84 \\ \hline
		LCVA & 55.63 & \textbf{0.794} \\ \hline
	\end{tabular}
	\vspace{1mm}
	\caption{Results for Amazon dataset-positive reviews.}
	\vspace{-1mm}
	\label{tab:amazon_pos}
	\begin{tabular}{ccc}
		\hline
		\textbf{Models} & \textbf{$\epsilon_{ATE}$} & \textbf{PEHE} \\ \hline
		OLS1 & 8.34 & 103.90 \\
		OLS2 & 7.99 & 92.27 \\
		RF & 9.46 & 83.92 \\
		CF & 13.49  & 153.35 \\
		CEVAE & 8.39 & 55.31 \\ \hline
		LVAE & \textbf{1.037} & \textbf{13.107} \\ \hline
	\end{tabular}
\end{table}

\vspace{-3mm}
\subsection{Results}
\vspace{-1mm}
Table~\ref{tab:ECON_result} compares the performance of LCVA along with other baselines for job training dataset, and Tables  ~\ref{tab:amazon_pos} and~\ref{tab:amazon_neg} reports the same for co-purchase dataset from Amazon. Overall, our model consistently outperforms the baselines in almost all scenarios. This observation can be explained by the fact that although models such as OLS1, OLS2 and random forest can learn the spillover effect, they only do so by controlling the observable features. Unfortunately, these features are inadequate to represent all the confounding variables. In comparison, LCVA learns representation for confounders with information extracted not only from features but also from treatments and factual outcomes. Another interesting observation is that for the job training dataset (Table~\ref{tab:ECON_result}), RF performs better than our model in terms of $\epsilon_{ATE}$. However, this scenario is different when it comes to Amazon dataset where LVAE is significantly better than RF on both positive and negative cases. A possible reason for this outcome could be attributed to the level (or intensity) of spillover effects in datasets. In job training dataset, we synthetically create the links between units through the K-NN graph, while in Amazon dataset, the link relationship between products is naturally present. This in turn implies that spillover effects are much stronger in Amazon dataset due the co-purchase behavior. Finally, it is also important to note that LCVA achieves better estimation of treatment effects (or counterfactual outcomes) when compared to the state-of-the-art CEVAE. This is because our model is specifically designed to capture the spillover effect between linked units.

\begin{table}[]
		\small
	\centering
	\caption{Results for Amazon dataset-negative reviews.}
	\label{tab:amazon_neg}
	\begin{tabular}{ccc}
		\hline
		\textbf{Models} & \textbf{$\epsilon_{ATE}$} & \textbf{PEHE} \\ \hline
		OLS1 & 11.25 & 52.50 \\
		OLS2 & 2.7 & 57.76 \\
		RF & 11.43 & 49.50 \\
		CF & 9.43 &  55.32 \\
		CEVAE & 7.64 & 43.72 \\ \hline
		LVAE & \textbf{1.218} & \textbf{13.107} \\ \hline
	\end{tabular}
\end{table}

\vspace{-3mm}
\section{Conclusion}
\vspace{-1mm}
In this paper, we propose a model called linked causal variational autoencoder (LCVA) that captures the spillover effect between pairs of units. Specifically, given a pair of units $u$ and $\bar{u}$, their individual treatment and outcomes, the encoder network of LCVA samples the confounders by conditioning on the observed covariates of u, the treatments of both $u$ and $\bar{u}$ and the outcome of $u$. Using a network of users from job training dataset (LaLonde (1986)) and co-purchase dataset from Amazon e-commerce domain, we show that LCVA is significantly more robust than existing methods in capturing spillover effects.
\vspace{-3mm}
\begin{acks}
	This material is based upon work supported by the National
	Science Foundation (NSF) grant 1614576, and the Office of
	Naval Research (ONR) grant N00014-17-1-2605.
\end{acks}
\vspace{-1mm}
\vspace{-2mm}
\bibliographystyle{ACM-Reference-Format}
\bibliography{CIKM18_2}
\end{document}